\documentclass[letterpaper]{article} % DO NOT CHANGE THIS
\usepackage[preprint]{aaai2027}  % DO NOT CHANGE THIS
\usepackage[hyphens]{url}  % DO NOT CHANGE THIS
\usepackage{graphicx} % DO NOT CHANGE THIS
\usepackage{natbib}  % DO NOT CHANGE THIS AND DO NOT ADD ANY OPTIONS TO IT
\usepackage{caption} % DO NOT CHANGE THIS AND DO NOT ADD ANY OPTIONS TO IT
\usepackage{booktabs}
\usepackage{amsmath}
\usepackage{arydshln}
\usepackage{colortbl}
\definecolor{tokS}{HTML}{EB6834}
\definecolor{vaeS}{HTML}{2A78D6}
\definecolor{ctlS}{HTML}{7D7B74}
\title{Where to Generate Touch:\\Do Image Latents Preserve Material Recognition?}
\author{\large
Wanghao~Ye\textsuperscript{1}\hspace{3pt}Aarosh~Das\textsuperscript{1}\hspace{3pt}Sihan~Chen\textsuperscript{2}\hspace{3pt}Yiting~Wang\textsuperscript{1}\hspace{3pt}Bowei~Tian\textsuperscript{1}\hspace{3pt}Guoheng~Sun\textsuperscript{1}\hspace{3pt}Shwai~He\textsuperscript{1}\\
Zheyu~Shen\textsuperscript{1}\hspace{3pt}Ziyao~Wang\textsuperscript{1}\hspace{3pt}Yexiao~He\textsuperscript{1}\hspace{3pt}Zhaoyi~Liu\textsuperscript{1}\hspace{3pt}Meng~Liu\textsuperscript{1}\hspace{3pt}Yuning~Zhang\textsuperscript{1}\hspace{3pt}Meng~Feng\textsuperscript{1}\hspace{3pt}Ziyi~Wang\textsuperscript{5}\\
Yilong~Dai\textsuperscript{3}\hspace{3pt}Siyuan~Peng\textsuperscript{1}\hspace{3pt}Zhenle~Duan\textsuperscript{1}\hspace{3pt}Joshua~Liu\textsuperscript{2}\hspace{3pt}Lang~Xiong\textsuperscript{4}\hspace{3pt}Daeun~Jung\textsuperscript{1}\\
Aditya~Sarkar\textsuperscript{1}\hspace{3pt}Parastoo~Pilevar\textsuperscript{1}\hspace{3pt}Xiaochuan~Gong\textsuperscript{1}\hspace{3pt}Yancheng~Zhu\textsuperscript{1}\hspace{3pt}Yingying~Fan\textsuperscript{1}\hspace{3pt}Qiong~Wang\textsuperscript{1}\hspace{3pt}Ang~Li\textsuperscript{1}\corresponding
}
\affiliations{
\textsuperscript{1}University of Maryland, College Park \quad \textsuperscript{2}Carnegie Mellon University \quad \textsuperscript{3}University of Alabama\\
\textsuperscript{4}Stanford University \quad \textsuperscript{5}Texas A\&M University \quad angliece@umd.edu
}

\newsavebox{\teaserbox}

\begin{document}

\maketitle
\sbox{\teaserbox}{\parbox[b]{\textwidth}{\centering
\includegraphics[width=\textwidth]{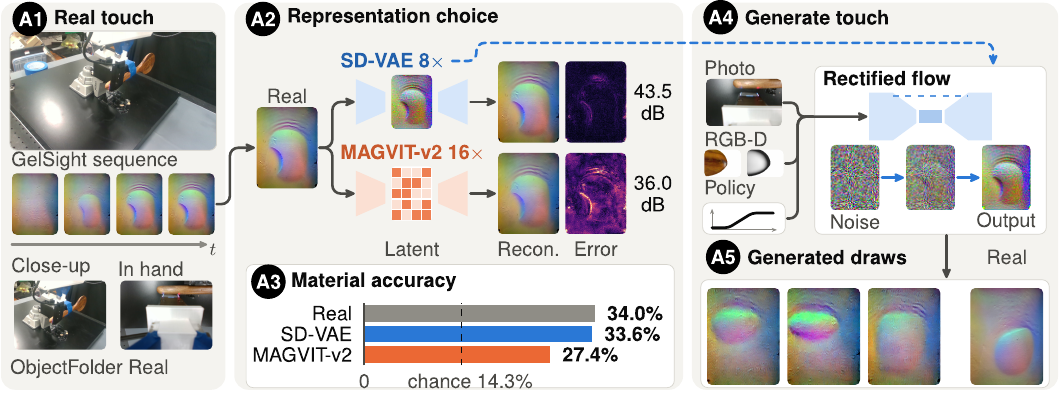}
\captionof{figure}{Choosing a representation and generating touch. A1 ObjectFolder Real recordings. A2 Continuous and discrete reconstructions with PSNR. A3 Real-trained material accuracy on 71 objects, family folds; drops in the text use unrounded accuracies. A4 Conditional generation in SD-VAE latents. A5 Three unselected draws beside real touch.}
\label{fig:system}}}
\begin{figure}[!b]
\makebox[\columnwidth][l]{\usebox{\teaserbox}}
\end{figure}
\begin{figure}[!b]
\rule{0pt}{\dimexpr\ht\teaserbox+\dp\teaserbox\relax}
\end{figure}

\begin{abstract}
Generated touch could extend robot learning beyond recorded contacts. A tactile generator needs a latent space, and we choose one by testing whether a material classifier trained on real GelSight presses still works on encode--decode reconstructions from pretrained image models. On ObjectFolder Real, the classifier loses 6.5 to 9.7 points on reconstructions from three 16$\times$ discrete tokenizers and 0.3 points on those from the Stable Diffusion VAE (SD-VAE). The 16$\times$ tokenizers again lose far more on Touch and Go and on DIGIT object recognition (YCB-Slide). Retraining the classifier on reconstructions, with no new labels, narrows the discrete gap. SD-VAE reconstructions also keep object identity, with 76.7\% object-matching accuracy against 78.9\% for real touch and 67.0\% for MAGVIT-v2. We therefore select SD-VAE latents and show a photo-to-touch generator trained in them. Code will be released upon acceptance.
\end{abstract}

\section{Introduction}
Tactile data are scarce because each recording requires physical contact \citep{wang2026vladatasurvey}. Generating touch from object photos could expand these data \citep{gungor2025leakage,ye2026tacgen}, provided that synthetic touch works with models trained on real recordings. GelSight and DIGIT capture elastomer deformation as RGB images \citep{yuan2017gelsight,lambeta2020digit}, so the continuous autoencoders \citep{rombach2022ldm} and discrete tokenizers \citep{esser2021vqgan,yu2024magvit2} of image foundation models can serve as the latent space of a tactile generator. We compare these spaces through encode--decode round trips, before any generator is trained. The comparison also shows how tokenizers from vision foundation models fare on a new sensory modality.

\begin{figure*}[t]
\centering
\includegraphics[width=\textwidth]{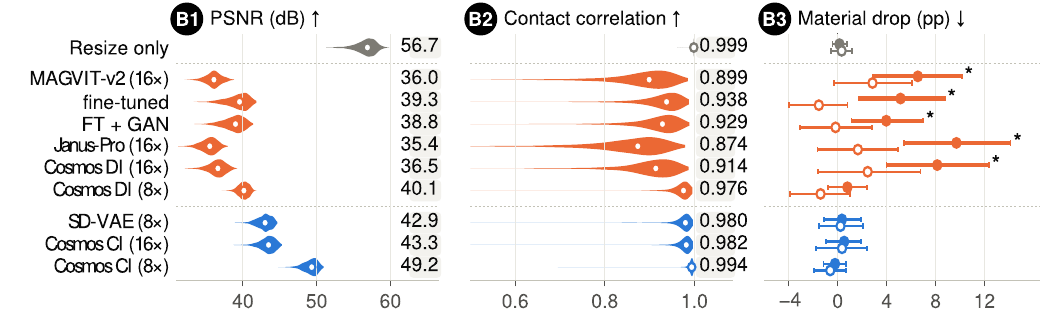}\par
\includegraphics[width=\textwidth]{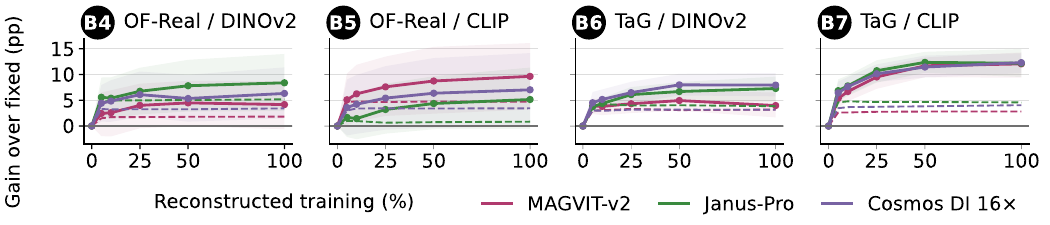}
\caption{Representation quality and probe recovery. B1--B3 Mean PSNR, median contact correlation and material drop on 2,093 OF-Real presses; B3 filled/open markers: real$\to$reconstruction/reverse transfer, 90\% family-bootstrap CIs ($*$: above zero). B4--B7 Probe retraining (solid) versus feature calibration (dashed), 95\% CIs; TaG: Touch and Go.}
\label{fig:fidelity}\label{fig:adaptation-dose}
\end{figure*}

Beyond pixel fidelity \citep{ye2026tacgen}, we test whether a material classifier trained on real touch still works on reconstructed presses. Continuous and discrete reconstructions both look close to the original (Figure~\ref{fig:system}, A2), yet across 71 objects the classifier loses 6.5 points on MAGVIT-v2 reconstructions and 0.3 on SD-VAE ones (A3; Figure~\ref{fig:fidelity}, B3). Retraining the classifier on reconstructions then shows how much of this loss is a mismatch between real and reconstructed inputs.

On two GelSight datasets, real-trained classifiers lose a quarter to a half of their margin over chance on reconstructions from released 16$\times$ discrete tokenizers and less than one point on SD-VAE reconstructions, and on DIGIT touch from YCB-Slide the discrete tokenizers lose 21 to 28 points of object-recognition accuracy against 1.2 for SD-VAE. SD-VAE reconstructions also retain high object-matching accuracy, and Section~\ref{sec:gen} shows a photo-to-touch generator trained in this compact space.

Our contributions are (1) a transfer test that ranks latent spaces for tactile generation with encode--decode passes and a lightweight classifier on frozen features; (2) a comparison of released and fine-tuned image tokenizers on three datasets from two sensor types, with grouped splits, several classifiers and bootstrap intervals; and (3) a matched comparison that separates quantization from downsampling, with classifier retraining that narrows the discrete gap without new labels.

\paragraph{Related work.} Tactile generators condition on local geometry \citep{bien2026flowtouch,bhouri2026multidiffsense}, on a sensor-aligned view of the contact region \citep{li2019connecting,yang2023generating,dou2024tarf} or on text \citep{tu2025texttoucher}. Recent tactile models generate with vector-quantized codes \citep{lyu2026vqtouch,huang2026htbench,tu2026unitac}, and TacGen predicts tactile latents from RGB \citep{ye2026tacgen}. Our study complements these generators (Appendix~\ref{app:numerical}). It compares candidate latent spaces on the same presses with a downstream classifier, under splits that keep product families apart \citep{gungor2025leakage}. Foundation models for other sensors face the same choice of input representation. Time-series models reprogram signals into language-model inputs \citep{jin2024timellm} or encode them as images \citep{wang2015imaging}, and language models answer questions about time series and reason over medical sensor data \citep{kong2025timemqa,langer2025opentslm}. Surveys and position papers cover Transformers, large models and reasoning for time series \citep{wen2023transformers,jin2023largemodels,jin2024position,kong2025position}.

\section{Measuring Material-Recognition Transfer}
\label{sec:test}

\paragraph{Test.} For reconstruction $R$ and a material classifier $f$ trained on real presses, we measure $\Delta_R=\mathrm{Acc}(f,x)-\mathrm{Acc}(f,R(x))$ on held-out presses $x$, following synthetic-image evaluation \citep{shmelkov2018good}. We also retrain on reconstructions and test on real presses \citep{esteban2017rcgan} and reconstructions to measure recovery. The test needs only encode--decode passes and a classifier on frozen features. We use this comparison to select a latent space that preserves material recognition before training a tactile generator.

\begin{figure*}[t]
\centering
\includegraphics[width=\textwidth]{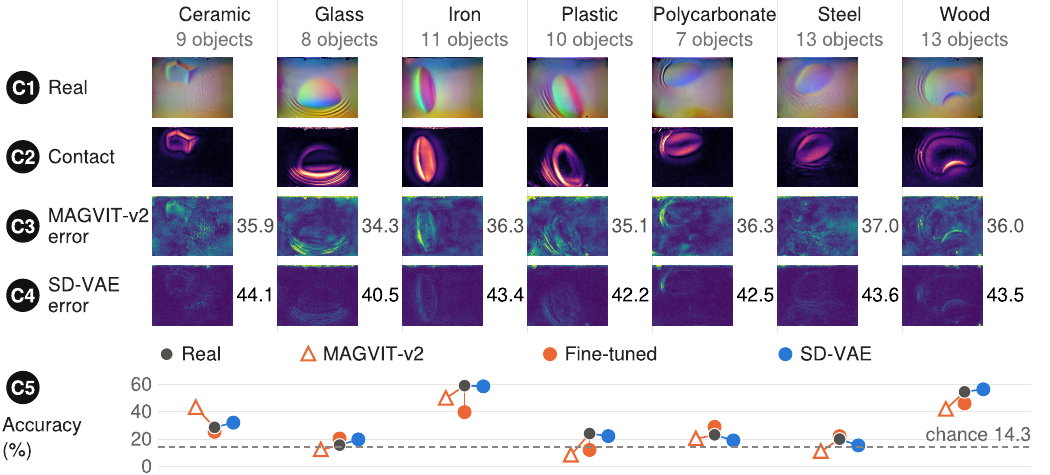}
\caption{Material cues and reconstruction errors. C1--C4 One training press per material: real last frame, contact map (absolute change from the first frame), MAGVIT-v2 and SD-VAE errors on a shared scale, with PSNR in dB. C5 Real-trained material accuracy on 71 objects, family folds.}
\label{fig:materials}
\end{figure*}

\paragraph{Data and splits.} We use the 71 training and validation objects of ObjectFolder Real (OF-Real), with 2,093 GelSight presses and seven materials \citep{gao2023objectfolder}; A1 shows the robot and the frames of one press. The dataset contains size and color variants of the same product, so each product family is kept within one of five cross-validation folds \citep{gungor2025leakage,kapoor2023leakage,roberts2017cv}. From a single press of an unseen product, the classifier identifies the material with 34.0\% accuracy, against 14.3\% chance. Object folds raise it to 38.9\% and the SD-VAE drop to 2.4 points; under every grouping the three 16$\times$ discrete drops are significant and larger (Appendix~\ref{app:folds}). The transfer tests follow two written protocols (Appendix~\ref{app:protocols}).

\paragraph{Classifiers and intervals.} The main classifier is a class-balanced logistic regression on frozen DINOv2-base features \citep{oquab2024dinov2}; classifiers on CLIP \citep{radford2021clip} and Sparsh \citep{higuera2024sparsh} features serve as checks. We call a drop significant when the lower end of its 90\% bootstrap interval over product families (recording days for Touch and Go) is above zero, a one-sided test at the 5\% level \citep{efron1994bootstrap}. Other comparisons use 95\% intervals.

\paragraph{Spaces.} The discrete candidates are the released MAGVIT-v2 and Janus-Pro tokenizers and the Cosmos DI tokenizers at 16$\times$ and 8$\times$ downsampling \citep{yu2024magvit2,chen2025januspro,sun2024llamagen,nvidia2025cosmos}. The continuous candidates are SD-VAE and the Cosmos CI tokenizers at 16$\times$ and 8$\times$. We also fine-tune MAGVIT-v2 with and without a GAN loss \citep{isola2017pix2pix}, retrain its decoder alone, and include resizing as a control. All round trips run at 512\,px; Appendix Table~\ref{tab:main} lists latent sizes and all drops.

\section{Transfer, Probe Retraining and Selection}
\label{sec:select}

\subsection{Released spaces}
Released 16$\times$ discrete tokenizers lower material accuracy by 6.5 (MAGVIT-v2), 9.7 (Janus-Pro) and 8.1 points (Cosmos), a third to a half of the classifier's margin over chance. They keep most of the contact geometry, with median contact-map correlations of 0.87 to 0.91, against 0.98 for SD-VAE. Pixel fidelity and contact geometry (B1,~B2) order the spaces much as the material drop does (Spearman $\rho=-0.95$ for PSNR). Among released spaces, the transfer test (B3) finds significant drops only for these three. Classifiers on CLIP and Sparsh features reproduce all three drops (Appendix~\ref{app:extra}).

The released MAGVIT-v2 loses most on plastic, wood, iron and steel and gains on ceramic; fine-tuning lowers iron accuracy by 19.4 points (C5). Across the material examples (\mbox{C1--C4}), reconstruction PSNR is 34 to 37\,dB for MAGVIT-v2 and 40 to 44\,dB for SD-VAE.

On 3,082 touches of 14 materials from Touch and Go \citep{yang2022touchandgo}, with folds by recording day, the pattern repeats. The three 16$\times$ discrete drops are 7.5, 8.4 and 11.8 points and the SD-VAE drop is 0.7. Cosmos DI 8$\times$ and CI 16$\times$ show small but significant drops of 2.0 and 1.5 points (Appendix~\ref{app:tag}).

\begin{figure*}[t]
\centering
\includegraphics[width=\textwidth]{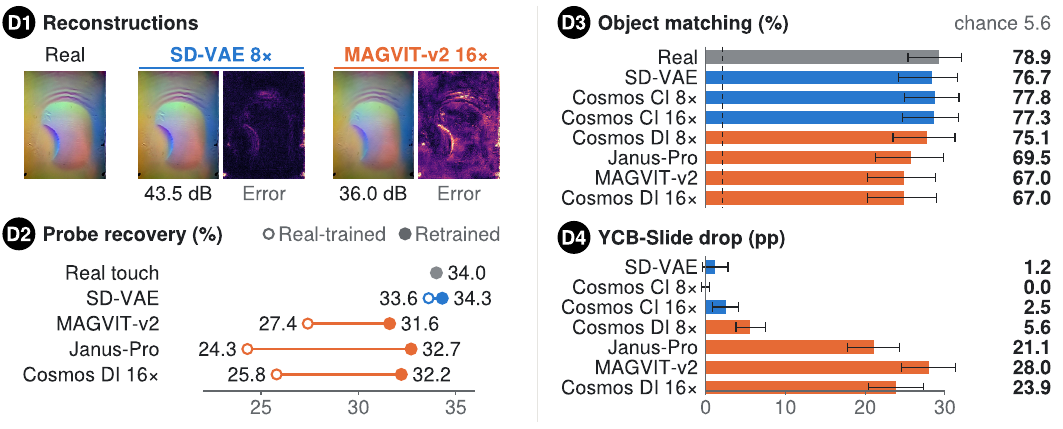}
\caption{Reconstruction, probe recovery and object-level tests. D1 Real touch, reconstructions and shared-scale errors. D2 Real-trained versus retrained classifiers on 71 objects, family folds, rounded after computing drops. D3 Object matching, all released spaces: 535 real queries, 18 objects, query site excluded; 95\% CIs. D4 YCB-Slide (DIGIT): drop in object-recognition accuracy, 3,000 touches, trajectory folds; 90\% CIs.}
\label{fig:representation-generation}
\label{fig:readout-controls}
\end{figure*}

We repeat the test with a second sensor type on YCB-Slide \citep{suresh2022midastouch}: 3,000 DIGIT frames from 50 sliding trajectories over 10 YCB objects, with object recognition as the task and folds by trajectory. The real-trained classifier reaches 77.3\% (chance 10\%). The three 16$\times$ discrete tokenizers lower it by 21.1 to 28.0 points and SD-VAE by 1.2 points (90\% CI [$-$0.4, 2.8]; D4). With contact-difference features alone, the drops are 25.2 to 31.0 points for the discrete tokenizers and 2.7 for SD-VAE (Appendix~\ref{app:ycb}).

PSNR does not settle close cases. On YCB-Slide, Cosmos CI 16$\times$ has as many latent values as SD-VAE and a higher PSNR (41.9 versus 40.7\,dB), yet its 2.5-point drop is significant and the 1.2 of SD-VAE is not; Cosmos DI 8$\times$ reaches 40.5\,dB and loses 5.6.

\subsection{Tokenizer fine-tuning and downsampling}
\label{sec:repair}
Fine-tuning MAGVIT-v2 on GelSight frames (39.3\,dB), adding a GAN loss, or retraining only its decoder (39.4\,dB) leaves significant drops of 4.0 to 5.1 points. The matched Cosmos pair separates quantization from downsampling. The discrete tokenizer loses 7.6 points more than its continuous counterpart at 16$\times$ and 1.0 more at 8$\times$; on Touch and Go the differences are 10.3 and 1.2 points. With the main classifier, most of the 16$\times$ loss therefore comes with quantization.

\subsection{Probe retraining}
\label{sec:adapt}
Retraining the classifier narrows the discrete gap with no new labels and no change to the tokenizer. Retraining replaces a share of the real training presses with their reconstructions and keeps their labels, so the label budget stays fixed. For the released 16$\times$ tokenizers on OF-Real, retrained classifiers reach 31.6 to 32.7\%, against 34.0\% on real touch (D2). On both datasets, replacing a quarter of the training presses gives most of the full-retraining gain (\mbox{B4--B7}), and on Touch and Go these gains are significant after multiplicity correction (Appendix Table~\ref{tab:followup-primary}).

\paragraph{Selection.} SD-VAE and Cosmos CI 8$\times$ show no significant drop on any dataset with the main classifier and primary splits, and the upper 90\% bounds of SD-VAE stay within a tenth of the margin over chance. SD-VAE uses four times fewer latent values ($64^2{\times}4$ versus $64^2{\times}16$), so we select it \citep{rombach2022ldm}. With CLIP on Touch and Go, SD-VAE loses 13.0 points, the 16$\times$ tokenizers 14.6 to 16.4 whether discrete or continuous, and Cosmos CI 8$\times$ 5.0 (Appendix Figure~\ref{fig:all-tag-clip}); when the readout is unknown, Cosmos CI 8$\times$ buys this robustness with four times more latent values.

\section{Object Identity and a Generation Example}
\label{sec:gen}
SD-VAE reconstructions also keep object identity. We match 535 real validation presses against galleries of the 18 validation objects, excluding each query's contact site (\mbox{Figure~\ref{fig:readout-controls}, D3}). Object-averaged accuracy is 78.9\% for real galleries, 76.7\% for SD-VAE reconstructions and 67.0\% for MAGVIT-v2 reconstructions (D1 shows one press). The SD-VAE gallery exceeds the MAGVIT-v2 gallery by 9.7 points (simultaneous 95\% CI [3.5, 16.6]; Appendix~\ref{app:figure-details}), and the Cosmos CI galleries reach 77.3\% and 77.8\%.

In the selected space, a conditional rectified-flow model \citep{liu2023rectifiedflow,lipman2023flowmatching} generates eight-frame presses from an object photo, contact-site geometry and a fixed pressing policy (A4). It is trained on 44 objects from other product families (Appendix~\ref{app:figure-details}), and A5 shows three unselected draws beside real touch.

\section{Conclusion and Broader Impact}
Our transfer test guides the choice of tactile generation space by measuring recognition accuracy on reconstructions. On three datasets and two sensor types, SD-VAE loses less than every released 16$\times$ discrete tokenizer, shows no significant drop under the main classifier and primary splits, and keeps object identity. With the main classifier the 16$\times$ loss comes mostly with quantization, and retraining the classifier narrows it without new labels. We therefore recommend SD-VAE latents for tactile generators used with real-trained models, and Cosmos CI 8$\times$ when the readout is unknown. The test can be run on any sensor whose signals are rendered as images, such as time series \citep{wang2015imaging}. Synthetic touch from photos and object geometry could let robots learn more from limited physical contact.
\label{end:main}

\bibliography{refs}
\appendix

\begin{table*}[t]
\centering
\small
\setlength{\tabcolsep}{1.35pt}
\newcommand{\ci}[2]{{\small[#1,\,#2]}}
\newcommand{\sw}[1]{{\color{#1}\rule[0.05ex]{1.3ex}{1.3ex}}\,}
\newcommand{\hi}{\cellcolor{tokS!22}}
\newcommand{\lat}[2]{{\small$#1^2{\times}$#2}}
\begin{tabular}{@{}lclll@{}}
\toprule
 & & \multicolumn{2}{c}{OF-Real drop (pts)\,$\downarrow$} & TaG \\
\cmidrule(lr){3-4}
Method & Latent & real$\to$rec. & rec.$\to$real & drop \\
\midrule
\rowcolor{black!4}\sw{ctlS}Resize only & -- & \phantom{$-$}0.1\,\ci{$-$0.4}{0.7} & \phantom{$-$}0.3\,\ci{$-$0.5}{1.1} & \phantom{1}0.6 \\
\cdashline{1-5}[0.6pt/1.5pt]
\multicolumn{5}{@{}l}{\rule{0pt}{2.6ex}\sw{tokS}\textit{Discrete, pretrained}} \\
\quad MAGVIT-v2 16$\times$ & \lat{32}{13\,b} & \hi\phantom{$-$}\textbf{6.5}\,\ci{2.9}{10.1} & \phantom{$-$}2.8\,\ci{$-$0.3}{6.1} & \hi\phantom{1}7.5 \\
\quad Janus-Pro 16$\times$ & \lat{32}{14\,b} & \hi\phantom{$-$}9.7\,\ci{5.4}{14.1} & \phantom{$-$}1.6\,\ci{$-$1.7}{4.9} & \hi\phantom{1}8.4 \\
\quad Cosmos DI 16$\times$ & \lat{32}{16\,b} & \hi\phantom{$-$}8.1\,\ci{4.0}{12.3} & \phantom{$-$}2.4\,\ci{$-$1.6}{6.8} & \hi 11.8 \\
\quad Cosmos DI 8$\times$ & \lat{64}{16\,b} & \phantom{$-$}0.8\,\ci{$-$0.8}{2.4} & $-$1.4\,\ci{$-$3.9}{1.0} & \hi\phantom{1}2.0 \\
\multicolumn{5}{@{}l}{\rule{0pt}{2.2ex}\sw{tokS}\textit{Discrete, fine-tuned} (MAGVIT-v2 16$\times$)} \\
\quad Fine-tuned & \lat{32}{13\,b} & \hi\phantom{$-$}5.1\,\ci{1.7}{8.7} & $-$1.5\,\ci{$-$4.0}{0.8} & \phantom{1}-- \\
\quad FT + GAN & \lat{32}{13\,b} & \hi\phantom{$-$}4.0\,\ci{1.1}{7.0} & $-$0.2\,\ci{$-$3.1}{2.8} & \phantom{1}-- \\
\midrule
\multicolumn{5}{@{}l}{\sw{vaeS}\textit{Continuous, pretrained}} \\
\quad SD-VAE 8$\times$ & \lat{64}{4\,f} & \phantom{$-$}0.3\,\ci{$-$1.1}{1.9} & \phantom{$-$}0.2\,\ci{$-$1.5}{2.1} & \phantom{1}0.7 \\
\quad Cosmos CI 16$\times$ & \lat{32}{16\,f} & \phantom{$-$}0.5\,\ci{$-$0.9}{1.9} & \phantom{$-$}0.3\,\ci{$-$1.8}{2.4} & \hi\phantom{1}1.5 \\
\quad Cosmos CI 8$\times$ & \lat{64}{16\,f} & $-$0.2\,\ci{$-$1.2}{0.7} & $-$0.6\,\ci{$-$1.9}{0.7} & \phantom{1}0.8 \\
\bottomrule
\end{tabular}
\caption{Material drop and latent size, family folds; $n\times$: spatial downsampling. Latent: positions $\times$ bits (b) or float channels (f) per position at 512\,px (Cosmos DI: 64,000 codes, about 16 bits). A drop is real-to-real accuracy (33.98\% on OF-Real; chance 14.3\%) minus the accuracy in the given direction, computed before rounding. TaG: real$\to$rec.\ drop in balanced accuracy on Touch and Go (Appendix~\ref{app:tag}), where the copies fine-tuned on OF-Real are not used. Brackets: 90\% CI; shaded: CI above zero. Fidelity: Figure~\ref{fig:fidelity}. FT: fine-tuned.}
\label{tab:main}
\end{table*}

\section{Fold Groupings}
\label{app:folds}

Real-to-real accuracy is 38.9\% with object folds and 34.0\% with family folds. As a sensitivity check, a broader grouping also joins matching sets of the same material and function, such as six wooden utensils; it gives a real-to-real accuracy of 27.4\%. Table~\ref{tab:folds} gives the real$\to$rec.\ drop of every method under all three groupings. We weight each object equally in all three groupings.

\begin{table*}[t]
\centering
\small
\setlength{\tabcolsep}{2.2pt}
\newcommand{\ci}[2]{{\small[#1,\,#2]}}
\newcommand{\hi}{\cellcolor{tokS!22}}
\begin{tabular}{@{}llll@{}}
\toprule
 & Object {\small(71)} & Family {\small(57)} & Broader {\small(39)} \\
\midrule
Resize only & \phantom{$-$}0.7\,\ci{$-$0.4}{1.8} & \phantom{$-$}0.1\,\ci{$-$0.4}{0.7} & \hi\phantom{$-$}0.9\,\ci{0.1}{1.8} \\
MAGVIT-v2 16$\times$ & \hi\phantom{$-$}6.8\,\ci{3.0}{10.9} & \hi\phantom{$-$}6.5\,\ci{2.9}{10.1} & \hi\phantom{$-$}4.6\,\ci{0.2}{8.8} \\
Janus-Pro 16$\times$ & \hi\phantom{$-$}10.3\,\ci{5.8}{15.0} & \hi\phantom{$-$}9.7\,\ci{5.4}{14.1} & \hi\phantom{$-$}7.0\,\ci{1.8}{11.9} \\
Cosmos DI 16$\times$ & \hi\phantom{$-$}6.4\,\ci{2.3}{10.5} & \hi\phantom{$-$}8.1\,\ci{4.0}{12.3} & \hi\phantom{$-$}7.0\,\ci{1.9}{12.0} \\
Cosmos DI 8$\times$ & \phantom{$-$}1.9\,\ci{$-$0.4}{4.6} & \phantom{$-$}0.8\,\ci{$-$0.8}{2.4} & \phantom{$-$}1.5\,\ci{$-$0.4}{3.3} \\
Fine-tuned & \hi\phantom{$-$}5.9\,\ci{2.3}{9.6} & \hi\phantom{$-$}5.1\,\ci{1.7}{8.7} & \hi\phantom{$-$}5.8\,\ci{1.6}{10.0} \\
FT + GAN & \hi\phantom{$-$}5.1\,\ci{1.8}{8.6} & \hi\phantom{$-$}4.0\,\ci{1.1}{7.0} & \hi\phantom{$-$}4.4\,\ci{0.9}{8.4} \\
SD-VAE 8$\times$ & \hi\phantom{$-$}2.4\,\ci{0.5}{4.5} & \phantom{$-$}0.3\,\ci{$-$1.1}{1.9} & \phantom{$-$}0.6\,\ci{$-$0.5}{1.6} \\
Cosmos CI 16$\times$ & \hi\phantom{$-$}2.2\,\ci{0.3}{4.2} & \phantom{$-$}0.5\,\ci{$-$0.9}{1.9} & \phantom{$-$}0.9\,\ci{$-$0.6}{2.3} \\
Cosmos CI 8$\times$ & \phantom{$-$}0.8\,\ci{$-$0.3}{1.9} & $-$0.2\,\ci{$-$1.2}{0.7} & \hi\phantom{$-$}0.6\,\ci{0.0}{1.3} \\
\bottomrule
\end{tabular}
\caption{Material drop (real$\to$rec., points) under three fold groupings on 71 objects, with resampled units in parentheses and 90\% CIs in brackets; shaded: CI above zero. A lower end shown as 0.0 is positive before rounding.}
\label{tab:folds}
\end{table*}

\section{Protocols and the 53-Object Subset}
\label{app:protocols}

Two written protocols fixed the transfer analyses before they were run. The first covered 53 training objects (1,557 presses) and five conditions, with the released MAGVIT-v2 drop under family folds as the primary comparison. The second added the Janus-Pro and Cosmos tokenizers and 18 validation objects, making the 71-object results primary. The five shared conditions retain their first-protocol results. On the 53-object subset, real-to-real accuracy is 39.8\% with object folds and 33.3\% with family folds. Under family folds the released MAGVIT-v2 drop is 4.8 points (90\% CI [0.9, 8.7]), the fine-tuned drops (3.0 and 3.2) are not significant, the SD-VAE drop is $-$0.3 (CI [$-$2.3, 1.6]), Janus-Pro drops 8.2 points ([3.1, 13.5]) and the Cosmos 16$\times$ discrete tokenizer 4.8 ([$-$0.9, 10.4]). The pooled discrete-minus-continuous Cosmos contrast is 2.2 ([$-$0.8, 5.3]).

\section{Implementation Details}
\label{app:implementation}

The probe uses DINOv2-base CLS features of the last frame and of the clipped difference image, plus the per-channel mean and standard deviation over frames, and a multinomial logistic regression with $C=1$, class-balanced weights and standardization fitted on the training folds only. Bootstrap CIs use 10,000 percentile replicates that resample objects (object folds) or fold units with their objects kept together (family folds). All reconstructions run at 512$\times$512 and are resized back. MAGVIT-v2 is the released \texttt{snu-aidas/magvitv2} checkpoint, the 8,192-code LFQ image tokenizer of Dynin-Omni. Cosmos tokenizers (Cosmos-0.1 CI/DI 8$\times$8 and 16$\times$16) run in bfloat16; the Janus-Pro tokenizer uses the generation-tokenizer weights of the released Janus-Pro-1B checkpoint; SD-VAE is sd-vae-ft-mse, decoded from the posterior mean. Each of two MAGVIT-v2 fine-tuned copies trains on half of the training objects and reconstructs the other half. Validation objects use a copy unexposed to their product family where available. One validation object (57) has no unexposed copy, which affects only the fine-tuned and decoder-retrained rows. The 18 test objects of OF-Real are held out for future work. Across the ten conditions of Table~\ref{tab:main}, mean PSNR (Figure~\ref{fig:fidelity}\,B1) and the DINOv2 drop have a Spearman correlation of $-$0.95.

\section{Further Probes and Decoder Retraining}
\label{app:extra}

\paragraph{Further probes.} Table~\ref{tab:probes} repeats the main comparison with two prespecified probes. The CLIP probe has the same construction as the DINOv2 probe, with CLIP ViT-B/16 image features. The touch probe applies the attentive head that Sparsh uses on its own benchmark to Sparsh features of the last two frames, each minus the pre-contact frame, and is trained per fold on real presses. A linear probe on the same Sparsh features reaches 21.5\%, but its CI lower bound of 18.6\% falls below the pre-specified threshold of five points above chance. We use the attentive head, which reaches 23.7\% with a lower bound of 20.6\% and meets this threshold. With both probes the drops of the three 16$\times$ discrete tokenizers are significant and the SD-VAE drop is not; across the three probes, these drops are 29 to 57\% of the probe's margin over chance. With CLIP, object folds show a small SD-VAE drop (3.4 points, CI [0.5, 6.4]), as with DINOv2, and the rec.$\to$real direction also drops for the Cosmos tokenizers, the continuous 16$\times$ one included. With DINOv2, the matched Cosmos discrete-minus-continuous gap is 7.6 points at 16$\times$ ([3.7, 11.6]) against 1.0 at 8$\times$ ([$-$0.3, 2.3]).

\begin{table*}[t]
\centering
\small
\setlength{\tabcolsep}{2.4pt}
\newcommand{\ci}[2]{{\small[#1,\,#2]}}
\newcommand{\hi}{\cellcolor{tokS!22}}
\begin{tabular}{@{}llll@{}}
\toprule
 & DINOv2 & CLIP & Sparsh \\
\midrule
Real$\to$real (\%) & 34.0 & 32.1 & 23.7 \\
MAGVIT-v2 16$\times$ & \hi 6.5\,\ci{2.9}{10.1} & \hi 10.1\,\ci{4.4}{15.9} & \hi 3.6\,\ci{1.1}{6.0} \\
Janus-Pro 16$\times$ & \hi 9.7\,\ci{5.4}{14.1} & \hi 5.1\,\ci{0.5}{9.9} & \hi 5.2\,\ci{1.4}{9.1} \\
Cosmos DI 16$\times$ & \hi 8.1\,\ci{4.0}{12.3} & \hi 8.1\,\ci{1.8}{14.1} & \hi 4.3\,\ci{1.4}{7.0} \\
SD-VAE 8$\times$ & 0.3\,\ci{$-$1.1}{1.9} & 1.7\,\ci{$-$3.1}{5.8} & 0.7\,\ci{$-$0.2}{1.6} \\
DI $-$ CI 16$\times$ & \hi 7.6\,\ci{3.7}{11.6} & \hi 7.2\,\ci{2.6}{11.8} & \hi 3.4\,\ci{0.4}{6.3} \\
$\rho$ with DINOv2 & -- & 0.83 & 0.87 \\
\bottomrule
\end{tabular}
\caption{Three probes: real$\to$rec.\ material drop (points) on 71 objects under family folds, with 90\% CIs; shaded: CI above zero. DI $-$ CI: discrete minus continuous Cosmos drop. Last row: Spearman correlation of the ten drops with the DINOv2 drops.}
\label{tab:probes}
\end{table*}

\paragraph{Decoder retraining.} We freeze MAGVIT-v2's encoder and quantizer, so every press keeps its released codes, and retrain only the decoder for 2,000 steps with an L1 and a VGG perceptual loss. As for the fine-tuned copies, each of two copies trains on one half of the training objects and reconstructs the other half. A second variant adds a loss on Sparsh features, computed on the same input as the touch probe. With family folds on 71 objects the drops are 4.7 (90\% CI [1.8, 7.7]) and 4.8 points ([1.9, 7.7]), against 6.5 for the released decoder, so most of the loss is already present in the codes; the recovery of 1.8 points has CIs [$-$1.5, 5.1] and [$-$1.7, 5.4]. PSNR rises to 39.4 and 37.0\,dB and median contact correlation to 0.94 and 0.95. With the touch-feature loss, a linear probe on Sparsh features no longer finds a drop ($-$1.3 points), while the DINOv2 probe still does.

\section{Training the Probe on Reconstructions}
\label{app:adaptation}
The reconstruction-to-reconstruction probe is fitted separately for each codec using reconstructed training presses, their material labels and training-fold standardization. It then classifies reconstructed presses from the held-out families. The DINOv2 features, regularization and family folds are identical to the real-trained comparison. Table~\ref{tab:adaptation} reports the results for the 71-object comparison. The tokenizer remains frozen during probe training. The reported remaining drops and their CIs compare each retrained probe with the real-to-real baseline. Table~\ref{tab:followup-primary} gives the paired recovery intervals for OF-Real, which has 57 product families.

\begin{table}[t]
\centering
\small
\setlength{\tabcolsep}{2.4pt}
\begin{tabular}{@{}lrrl@{}}
\toprule
 & \multicolumn{2}{c}{Training presses} & Remaining \\
\cmidrule(lr){2-3}
Method & Real & Recon. & drop (pts) \\
\midrule
Resize only & 33.8 & 34.0 & 0.0 {\small[$-$0.7, 0.7]} \\
MAGVIT-v2 16$\times$ & 27.4 & 31.6 & 2.4 {\small[0.6, 4.2]} \\
Janus-Pro 16$\times$ & 24.3 & 32.7 & 1.3 {\small[$-$1.1, 3.8]} \\
Cosmos DI 16$\times$ & 25.8 & 32.2 & 1.8 {\small[$-$0.1, 3.6]} \\
Cosmos DI 8$\times$ & 33.2 & 34.8 & $-$0.8 {\small[$-$2.4, 0.7]} \\
Fine-tuned & 28.8 & 33.9 & 0.1 {\small[$-$1.7, 2.0]} \\
FT + GAN & 30.0 & 33.3 & 0.7 {\small[$-$0.9, 2.2]} \\
SD-VAE 8$\times$ & 33.6 & 34.3 & $-$0.3 {\small[$-$1.6, 0.8]} \\
Cosmos CI 16$\times$ & 33.5 & 34.3 & $-$0.3 {\small[$-$2.0, 1.2]} \\
Cosmos CI 8$\times$ & 34.2 & 34.0 & $-$0.1 {\small[$-$0.9, 0.8]} \\
\bottomrule
\end{tabular}
\caption{Material accuracy on reconstructed test presses (\%), with real or reconstructed training presses. The real-to-real baseline is 34.0\%; chance is 14.3\%. Last column: baseline minus reconstruction-to-reconstruction accuracy, with 90\% family-bootstrap CIs. All 71 objects use the same held-out family folds as Table~\ref{tab:main}.}
\label{tab:adaptation}
\end{table}

\paragraph{CLIP check.} The CLIP probe uses the same folds and training construction. Its real-to-real accuracy is 32.1\%. Training on reconstructions raises the released MAGVIT-v2, Janus-Pro and Cosmos DI 16$\times$ accuracies from 22.0\%, 27.0\% and 24.0\% to 31.7\%, 32.1\% and 31.1\%, respectively. Table~\ref{tab:clip-adaptation} reports the CLIP results. Appendix~\ref{app:dose} extends this analysis to Touch and Go with both feature extractors.

\begin{table}[t]
\centering
\small
\setlength{\tabcolsep}{2.4pt}
\begin{tabular}{@{}lrrl@{}}
\toprule
 & \multicolumn{2}{c}{Training presses} & Remaining \\
\cmidrule(lr){2-3}
Method & Real & Recon. & drop (pts) \\
\midrule
Resize only & 32.1 & 32.3 & $-$0.2 {\small[$-$1.1, 0.6]} \\
MAGVIT-v2 16$\times$ & 22.0 & 31.7 & 0.4 {\small[$-$1.4, 2.3]} \\
Janus-Pro 16$\times$ & 27.0 & 32.1 & 0.0 {\small[$-$2.0, 2.0]} \\
Cosmos DI 16$\times$ & 24.0 & 31.1 & 1.0 {\small[$-$1.2, 3.3]} \\
Cosmos DI 8$\times$ & 32.3 & 33.1 & $-$1.0 {\small[$-$2.8, 0.7]} \\
Fine-tuned & 24.5 & 31.8 & 0.3 {\small[$-$1.8, 2.4]} \\
FT + GAN & 29.4 & 30.9 & 1.2 {\small[$-$1.5, 4.2]} \\
SD-VAE 8$\times$ & 30.4 & 32.4 & $-$0.3 {\small[$-$2.3, 1.9]} \\
Cosmos CI 16$\times$ & 31.3 & 31.0 & 1.1 {\small[$-$0.6, 2.8]} \\
Cosmos CI 8$\times$ & 32.5 & 32.2 & $-$0.1 {\small[$-$1.4, 1.2]} \\
\bottomrule
\end{tabular}
\caption{CLIP material accuracy on reconstructed presses (\%), under the same family folds. Last column: real-to-real baseline (32.1\%) minus reconstruction-to-reconstruction accuracy, with 90\% family-bootstrap CIs.}
\label{tab:clip-adaptation}
\end{table}

\section{Touch and Go}
\label{app:tag}

Touch and Go \citep{yang2022touchandgo} records GelSight presses on objects in natural scenes, with a material label for every frame inside a press. We take each run of labelled frames as one touch, keep the 14 materials with at least 80 touches on at least 10 recording days, and sample up to 300 touches per material, 3,082 in total. Each touch contributes its middle labelled frame and the frame ten before its onset, which play the roles of the last and first frames of an OF-Real press. Folds and bootstrap units are the 24 recording days. We report class-balanced accuracy (chance 7.1\%; real-to-real 35.2\%). The fine-tuned MAGVIT-v2 copies were trained on OF-Real and are not used. Unlike on OF-Real, the 16$\times$ discrete tokenizers also drop in the rec.$\to$real direction, and the 8$\times$ discrete and 16$\times$ continuous Cosmos tokenizers drop slightly (Table~\ref{tab:tag}).

\begin{table*}[t]
\centering
\small
\setlength{\tabcolsep}{1.9pt}
\newcommand{\ci}[2]{{\small[#1,\,#2]}}
\newcommand{\hi}{\cellcolor{tokS!22}}
\begin{tabular}{@{}lccll@{}}
\toprule
Method & PSNR & $r_c$ & real$\to$rec. & rec.$\to$real \\
\midrule
Resize only & 47.3 & 0.999 & \phantom{$-$}0.6\,\ci{$-$0.1}{1.2} & $-$0.4\,\ci{$-$1.1}{0.4} \\
MAGVIT-v2 16$\times$ & 29.4 & 0.936 & \hi\phantom{$-$}7.5\,\ci{6.4}{8.9} & \hi\phantom{$-$}8.1\,\ci{5.8}{10.3} \\
Janus-Pro 16$\times$ & 29.0 & 0.938 & \hi\phantom{$-$}8.4\,\ci{6.9}{9.7} & \hi\phantom{$-$}8.5\,\ci{5.7}{11.0} \\
Cosmos DI 16$\times$ & 29.0 & 0.932 & \hi\phantom{$-$}11.8\,\ci{10.1}{13.6} & \hi\phantom{$-$}13.8\,\ci{11.2}{15.7} \\
Cosmos DI 8$\times$ & 35.2 & 0.986 & \hi\phantom{$-$}2.0\,\ci{0.8}{3.1} & \phantom{$-$}1.1\,\ci{$-$0.4}{2.5} \\
SD-VAE 8$\times$ & 37.3 & 0.990 & \phantom{$-$}0.7\,\ci{$-$0.7}{1.8} & $-$0.2\,\ci{$-$1.5}{1.2} \\
Cosmos CI 16$\times$ & 36.3 & 0.989 & \hi\phantom{$-$}1.5\,\ci{0.0}{2.9} & \hi\phantom{$-$}2.0\,\ci{0.4}{3.6} \\
Cosmos CI 8$\times$ & 42.4 & 0.997 & \phantom{$-$}0.8\,\ci{$-$0.2}{1.6} & $-$0.4\,\ci{$-$1.3}{0.6} \\
\bottomrule
\end{tabular}
\caption{Touch and Go: fidelity and drop in balanced material accuracy (points) over 3,082 touches, folds and bootstrap by recording day; brackets: 90\% CI; shaded: CI above zero (a lower end shown as 0.0 is positive before rounding). Discrete minus continuous Cosmos: 10.3 points at 16$\times$ (90\% CI [8.1, 12.6]), 1.2 at 8$\times$ ([0.3, 2.2]).}
\label{tab:tag}
\end{table*}

\section{YCB-Slide}
\label{app:ycb}
YCB-Slide \citep{suresh2022midastouch} contains DIGIT recordings of a sensor slid over 10 YCB objects, five trajectories per object. From each trajectory we take 60 frames at evenly spaced indices between 5\% and 95\% of its length, 3,000 touches in total, without inspecting or filtering frames. The reference frame of a trajectory is the DIGIT background image closest to the median of its 60 frames. The task is 10-way object recognition with the classifier and bootstrap of the Touch and Go analysis. Folds and bootstrap clusters are trajectories, so each object is tested on a trajectory absent from training. The protocol, including a secondary analysis on the contact-difference features alone, was written before any frame was opened. Table~\ref{tab:ycb} lists every condition.

\begin{table*}[t]
\centering
\small
\setlength{\tabcolsep}{4pt}
\begin{tabular}{@{}lrrll@{}}
\toprule
Reconstruction & PSNR (dB) & Contact $r$ & Drop, full features & Drop, difference only \\
\midrule
Resize only & 60.3 & 0.999 & $-$0.1 {\small[$-$0.5, 0.3]} & 0.1 {\small[$-$0.2, 0.5]} \\
SD-VAE (8$\times$) & 40.7 & 0.976 & 1.2 {\small[$-$0.4, 2.8]} & 2.7 {\small[1.0, 4.5]} \\
Cosmos CI 8$\times$ & 50.0 & 0.995 & 0.0 {\small[$-$0.5, 0.5]} & 0.3 {\small[$-$0.2, 0.7]} \\
Cosmos CI 16$\times$ & 41.9 & 0.977 & 2.5 {\small[0.9, 4.1]} & 1.8 {\small[0.4, 3.2]} \\
Cosmos DI 8$\times$ & 40.5 & 0.975 & 5.6 {\small[3.8, 7.5]} & 7.5 {\small[5.2, 9.7]} \\
Janus-Pro (16$\times$) & 32.2 & 0.917 & 21.1 {\small[17.8, 24.3]} & 25.2 {\small[21.7, 28.7]} \\
MAGVIT-v2 (16$\times$) & 34.5 & 0.937 & 28.0 {\small[24.6, 31.3]} & 31.0 {\small[27.7, 34.3]} \\
Cosmos DI 16$\times$ & 36.2 & 0.934 & 23.9 {\small[20.4, 27.3]} & 25.5 {\small[22.3, 28.7]} \\
\bottomrule
\end{tabular}
\caption{YCB-Slide (DIGIT): reconstruction fidelity and drop in balanced object-recognition accuracy (points) over 3,000 touches from 50 trajectories, with 90\% trajectory-bootstrap CIs. Real-to-real accuracy is 77.3\% with full features and 71.2\% with difference features alone; chance is 10\%.}
\label{tab:ycb}
\end{table*}

\section{Per-Material Breakdown}
\label{app:materials}

Figure~\ref{fig:materials}\,C5 breaks the probe accuracy down by material. Each material has seven to thirteen objects. The released MAGVIT-v2 lowers accuracy on plastic, wood, iron and steel and raises it on ceramic; fine-tuning hurts iron most.

\section{Further Related Work}
\label{app:numerical}

FlowTouch generates tactile images from an object's local mesh \citep{bien2026flowtouch}, MultiDiffSense from depth and contact pose \citep{bhouri2026multidiffsense}, VQ-Touch and HT-Bench with vector-quantized codes \citep{lyu2026vqtouch,huang2026htbench}, and UniTac within one model for tactile understanding and generation \citep{tu2026unitac}. Vision-to-touch generators mostly condition on a sensor-aligned view of the contact region \citep{li2019connecting,yang2023generating,dou2024tarf,yang2022touchandgo} or on text that describes the material \citep{tu2025texttoucher}, and TacGen predicts tactile latents from RGB \citep{ye2026tacgen}. We compare the representations used for tactile generation and then train a conditional model in the selected space. Touch--language models usually pass features from a dedicated tactile encoder to the language model rather than tokenizing tactile images \citep{yang2024unitouch,fu2024tvl,yu2024octopi}.

\begin{table}[t]
\centering
\small
\setlength{\tabcolsep}{3pt}
\begin{tabular}{@{}lll@{}}
\toprule
Gallery & Full features & Difference only \\
\midrule
Real & 78.9 {\small[68.5, 86.6]} & 66.1 {\small[55.6, 75.6]} \\
Resize only & 78.3 {\small[67.9, 86.3]} & 65.9 {\small[55.1, 75.5]} \\
SD-VAE & 76.7 {\small[65.4, 85.4]} & 62.9 {\small[51.9, 73.0]} \\
MAGVIT-v2 & 67.0 {\small[54.9, 77.7]} & 53.9 {\small[39.6, 67.7]} \\
Janus-Pro & 69.5 {\small[57.4, 80.4]} & 52.4 {\small[38.4, 66.4]} \\
Cosmos DI 16$\times$ & 67.0 {\small[54.8, 78.1]} & 49.7 {\small[36.4, 63.1]} \\
Cosmos DI 8$\times$ & 75.1 {\small[63.4, 84.4]} & 61.4 {\small[50.0, 72.1]} \\
Cosmos CI 16$\times$ & 77.3 {\small[66.7, 85.5]} & 64.0 {\small[53.2, 74.0]} \\
Cosmos CI 8$\times$ & 77.8 {\small[67.3, 85.8]} & 64.1 {\small[53.9, 73.0]} \\
\bottomrule
\end{tabular}
\caption{Object matching on 535 queries from 18 objects. Accuracy (\%) with 95\% object-bootstrap CIs; chance is 5.6\%.}
\label{tab:codec-matching}
\end{table}

\begin{table*}[t]
\centering\small
\setlength{\tabcolsep}{5pt}
\begin{tabular}{@{}llrrr@{}}
\toprule
Dataset / feature & Codec & 25\% refit $-$ fixed & 25\% refit $-$ calibration & Full refit $-$ fixed \\
\midrule
OF-Real / DINOv2 & MAGVIT-v2 & 4.0 {\small[-3.1, 10.6]} & 2.3 {\small[-0.4, 5.3]} & 4.2 {\small[-3.1, 10.8]} \\
 & Janus-Pro & 6.8 {\small[-0.4, 14.0]} & 1.8 {\small[-0.9, 4.5]} & 8.4 {\small[0.1, 16.6]} \\
 & Cosmos DI 16$\times$ & 6.1 {\small[-0.7, 13.5]} & 2.8 {\small[0.02, 6.2]} & 6.3 {\small[-0.7, 13.9]} \\
\addlinespace
OF-Real / CLIP & MAGVIT-v2 & 7.6 {\small[-1.3, 16.8]} & 3.0 {\small[-0.6, 6.8]} & 9.7 {\small[0.3, 19.1]} \\
 & Janus-Pro & 3.2 {\small[-4.0, 11.3]} & 2.6 {\small[-1.3, 6.8]} & 5.2 {\small[-2.9, 14.4]} \\
 & Cosmos DI 16$\times$ & 5.4 {\small[-4.3, 15.3]} & 2.0 {\small[-1.3, 5.8]} & 7.0 {\small[-3.5, 17.5]} \\
\addlinespace
Touch and Go / DINOv2 & MAGVIT-v2 & 4.3 {\small[1.3, 7.1]} & 1.0 {\small[-1.7, 3.6]} & 4.0 {\small[0.7, 7.6]} \\
 & Janus-Pro & 6.1 {\small[2.9, 8.5]} & 2.1 {\small[-0.03, 4.6]} & 7.3 {\small[3.4, 10.7]} \\
 & Cosmos DI 16$\times$ & 6.5 {\small[3.3, 10.0]} & 3.3 {\small[1.0, 6.4]} & 8.0 {\small[4.5, 11.2]} \\
\addlinespace
Touch and Go / CLIP & MAGVIT-v2 & 9.5 {\small[6.0, 12.5]} & 6.8 {\small[3.7, 9.3]} & 12.1 {\small[8.2, 15.5]} \\
 & Janus-Pro & 10.8 {\small[7.1, 14.1]} & 6.1 {\small[2.9, 9.2]} & 12.2 {\small[8.4, 16.1]} \\
 & Cosmos DI 16$\times$ & 10.1 {\small[7.6, 13.4]} & 6.4 {\small[4.0, 9.1]} & 12.3 {\small[9.3, 15.0]} \\
\bottomrule
\end{tabular}
\caption{Primary probe-retraining contrasts (points). Brackets are simultaneous 95\% Bonferroni intervals over 24 comparisons for the two 25\% columns, and over 12 comparisons for the full-retraining column. OF-Real uses object-equal accuracy and family resampling; Touch and Go uses balanced accuracy and day resampling. Endpoints near zero keep two decimals.}
\label{tab:followup-primary}
\end{table*}

\begin{table*}[t]
\centering\small
\setlength{\tabcolsep}{4pt}
\begin{tabular}{@{}lrrr@{}}
\toprule
Readout / contrast & MAGVIT-v2 & Janus-Pro & Cosmos DI 16$\times$ \\
\midrule
Fixed-probe accuracy & 0.0 {\small[0.0, 0.0]} & 0.0 {\small[0.0, 0.0]} & 0.3 {\small[-0.9, 1.3]} \\
25\% refit accuracy & 0.01 {\small[-0.03, 0.1]} & 0.0 {\small[-0.04, 0.03]} & 0.1 {\small[-0.4, 0.9]} \\
25\% calibration accuracy & 0.0 {\small[0.0, 0.0]} & 0.0 {\small[0.0, 0.0]} & 0.2 {\small[-0.5, 1.0]} \\
Full-refit accuracy & -0.02 {\small[-0.1, 0.1]} & 0.03 {\small[0.0, 0.1]} & 0.2 {\small[-1.1, 1.7]} \\
25\% refit $-$ fixed gain & 0.01 {\small[-0.03, 0.1]} & 0.00 {\small[-0.04, 0.03]} & -0.1 {\small[-1.1, 1.2]} \\
Full refit $-$ fixed gain & -0.02 {\small[-0.1, 0.1]} & 0.03 {\small[0.0, 0.1]} & -0.05 {\small[-1.5, 1.5]} \\
\bottomrule
\end{tabular}
\caption{Numerical sensitivity on Touch and Go: new paired-extraction DINOv2 results minus the original cached-feature DINOv2 results, in percentage points, with paired pointwise 95\% day-bootstrap intervals. The last two rows give changes in recovery gains. The paired extraction computes DINOv2 and CLIP features from the same reconstructed images.}
\label{tab:numerical-audit}
\end{table*}

\section{Generator and Object Matching}
\label{app:figure-details}
The robot photos and tactile frames in Figure~\ref{fig:system}\,A1 are from ObjectFolder Real \citep{gao2023objectfolder}. The reconstruction example is training press 11:0:0. Sensor images are contrast-stretched, pixel-difference maps share a scale, and PSNR is measured on the last frame. Material-drop estimates use 71 objects under product-family folds, with 90\% CIs.

Generated examples show the first validation object by ID (15) at site 17, with three unselected draws from the final checkpoint. Their last frames and the real press share a contrast stretch: deviations from the pooled per-channel mean are multiplied by 1.6 and recentered at mid-grey.

\paragraph{Generator and training.} The model predicts eight evenly spaced frames of each press, at normalized times $\{0,57,\dots,399\}/399$. A frozen SD-VAE encodes each frame's posterior mean at 512$\times$512 into four 64$\times$64 latent channels, scaled by 0.18215. The velocity network is a sequence U-Net with base width 32, temporal mixing and joint bottleneck attention. Spatial geometry channels are concatenated to noisy latents; the photo embedding, pressing policy and frame time condition the network. Geometry is a six-channel rendering of the contact region; the orientation transform is selected using training presses only. Every press uses the same policy.

Training minimizes the squared rectified-flow velocity error between Gaussian noise and the tactile latent, with uniformly sampled interpolation times. We use AdamW (learning rate $2\times10^{-4}$, weight decay 0.01), object-balanced batches of 16, gradient clipping at 1 and fp32 computation. We train for 30,000 updates and use the raw weights of the final checkpoint. Sampling uses 25 Euler steps with a fixed seed per query.

\paragraph{Object-gallery diagnostic.} Object matching uses 535 real query presses from 18 validation objects and the prespecified full-feature centroid score. Each query's site is excluded from the gallery. The real gallery uses other real presses. Features concatenate DINOv2 last-frame and contact-difference embeddings with six RGB statistics (1,542 dimensions). They are standardized using 1,353 real presses from the 44 training objects and then L2-normalized. Each object's remaining gallery features are averaged and normalized before cosine matching. Accuracy averages presses within objects, then weights the 18 objects equally.

The reconstruction galleries were added to this prespecified diagnostic later, and Table~\ref{tab:codec-matching} reports every released codec. The difference-only check uses the 768 contact-difference features, omitting the last-frame embedding and global RGB statistics. CIs resample the 18 objects jointly across conditions (10,000 replicates, seed 20261108). Specified contrasts use simultaneous 95\% Bonferroni intervals over three comparisons. The full-feature real-minus-SD-VAE difference is 2.2 points (CI [$-$0.4, 5.8]); the difference-only SD-VAE-minus-MAGVIT-v2 contrast is 9.0 points ([1.2, 17.6]). With all features, the SD-VAE-minus-MAGVIT-v2 contrast is 9.7 points ([3.5, 16.6]).

\section{Retraining under a Fixed Label Budget}
\label{app:dose}

We specified this exposure study after the transfer and full-retraining results, fixing the doses, sampling seeds, controls and contrast families before computing the curves. It covers the released reconstruction conditions on OF-Real and Touch and Go with DINOv2 and CLIP, using the same outer folds as the transfer tests. Encoders remain frozen. OF-Real uses object-equal accuracy with product-family folds; Touch and Go uses balanced accuracy with recording-day folds. Its CLIP probe uses the same reference and contact frames as its DINOv2 probe.

\paragraph{Exposure and labels.} Within each outer training fold, we select 0\%, 5\%, 10\%, 25\%, 50\% or 100\% of the training examples for reconstruction-domain exposure. For each material, examples are ordered by a hash of their identifier and one of three fixed sampling seeds; a positive dose selects the first $\lceil d n_c\rceil$ examples. Subsets are nested and shared across codecs and feature extractors. Each selected example's real feature is replaced by its reconstructed feature. This one-for-one replacement keeps the number of distinct labeled examples constant. The dose is the fraction of training examples replaced with reconstructions.

\paragraph{Refitting and calibration.} At each dose we fit the original class-balanced multinomial logistic regression from zero, using $C=1$, 300 L-BFGS iterations and training-only standardization. Zero exposure recovers the fixed real-trained probe; full exposure recovers reconstruction-to-reconstruction training. As a control, we retain the original classifier and its scaler, estimate coordinate-wise means and population standard deviations from the selected paired real and reconstructed training features, and transform each reconstructed test feature as
\[
\widetilde z=\frac{z-\mu_R}{\sigma_R+10^{-6}}\,(\sigma_X+10^{-6})+\mu_X.
\]
The arithmetic is coordinate-wise. Calibration statistics come only from paired training features. At zero dose, the control uses the original fixed probe. Refitting gains are measured against this diagonal mean/variance correction.

\paragraph{Uncertainty and prespecified comparisons.} We average each test example's correctness across the three sampling seeds before computing the metric. Paired bootstrap intervals use 20,000 shared draws of product families for OF-Real or recording days for Touch and Go. Family resampling preserves object-equal weighting; day resampling recomputes class-balanced accuracy. The intervals condition on the fitted models and training data; curve intervals are two-sided and pointwise at 95\%.

The primary dose is 25\%. For each dataset, feature extractor and released 16$\times$ discrete codec, we compare refitting with fixed transfer and with the statistics-only control. These 24 contrasts additionally receive simultaneous 95\% Bonferroni percentile intervals. Full-retraining gains form a separate family of 12 contrasts with the same correction. Table~\ref{tab:followup-primary} reports these contrasts. The transfer tests use 90\% intervals.

\paragraph{Numerical pairing and sensitivity.} The primary OF-Real and Touch and Go DINOv2 analyses use saved features. For the Touch and Go CLIP analysis, we extract DINOv2 and CLIP features from the same reconstructed images, with identical image-statistic coordinates. On fixed reference inputs the other codecs reproduce the saved features exactly, while Cosmos decoding on the GPU is not bit-exact.

We specified a sensitivity analysis at 0\%, 25\% and 100\% exposure for the three primary codecs before computing the full Touch and Go CLIP results. It applies the original DINOv2 source classifiers and the same folds, real training features, selected subsets and fitting procedure to the newly paired reconstructions. Table~\ref{tab:numerical-audit} reports changes in accuracy and recovery gains with paired pointwise 95\% day-bootstrap intervals. The primary DINOv2 contrasts in Table~\ref{tab:followup-primary} use the original cached features.

For Cosmos DI 16$\times$, the fixed classifier changes its predicted label on 438 of 3,082 touches. Its balanced accuracy changes by 0.26 points (95\% CI [$-$0.90, 1.25]). The 25\% and full-retraining recovery gains change by $-$0.12 ([$-$1.13, 1.22]) and $-$0.05 points ([$-$1.54, 1.49]), respectively. The full-gain estimate is 7.90 points with the paired extraction, versus 7.95 with the original cache. MAGVIT-v2 and Janus-Pro full-gain estimates change by $-$0.02 and 0.03 points.

\paragraph{Exposure-level summary.} On Touch and Go, gains at 25\% exposure are positive after multiplicity correction for all three codecs with both feature extractors. At 25\% exposure, only Cosmos DI 16$\times$ clearly exceeds calibration with DINOv2, whereas all three CLIP refits do, by 6.1--6.8 points (Table~\ref{tab:followup-primary}).

\section{Complete Probe-Retraining Curves}
\label{app:all-curves}
\begin{figure*}[p]
\centering
\includegraphics[width=0.83\textwidth]{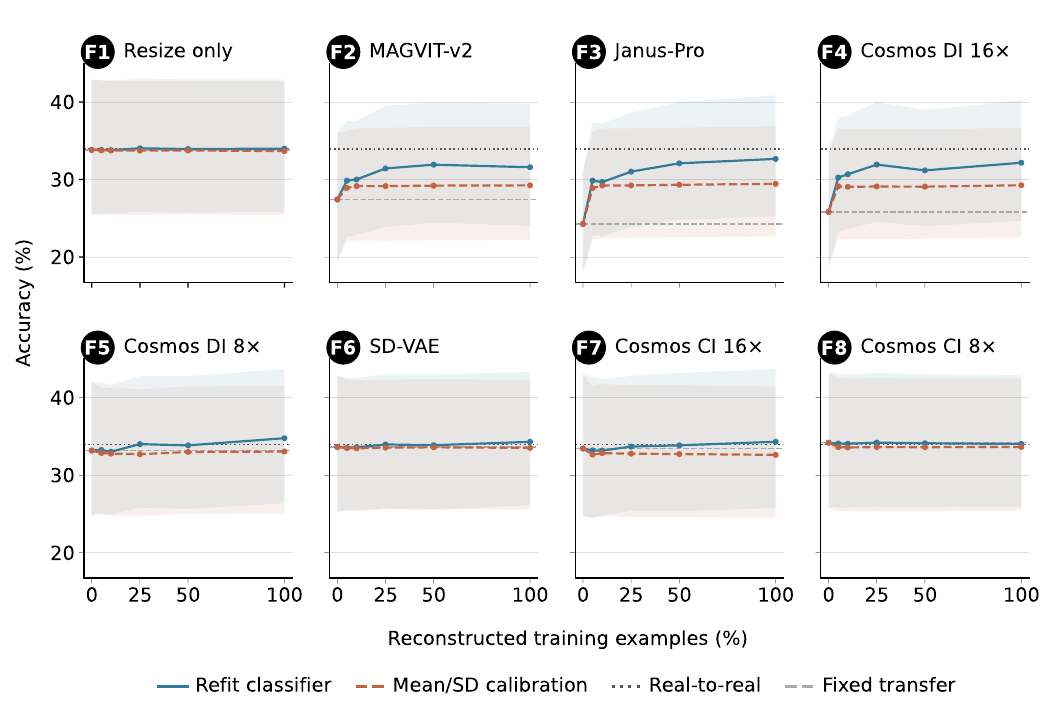}
\caption{\mbox{F1--F8} OF-Real / DINOv2. Every reconstruction condition, with paired pointwise 95\% cluster-bootstrap intervals. The horizontal references are real-to-real accuracy (dotted) and fixed-probe transfer (grey dashed).}
\label{fig:all-of-real-dino}
\includegraphics[width=0.83\textwidth]{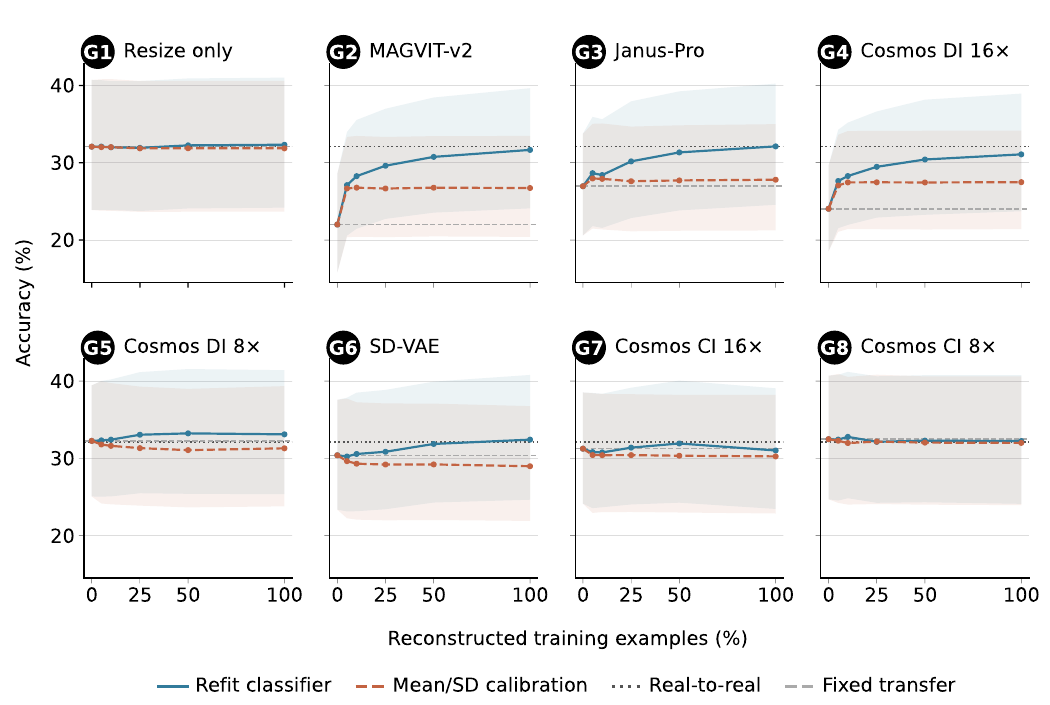}
\caption{\mbox{G1--G8} OF-Real / CLIP. Every reconstruction condition, with paired pointwise 95\% cluster-bootstrap intervals. The horizontal references are real-to-real accuracy (dotted) and fixed-probe transfer (grey dashed).}
\label{fig:all-of-real-clip}
\end{figure*}

\begin{figure*}[p]
\centering
\includegraphics[width=0.83\textwidth]{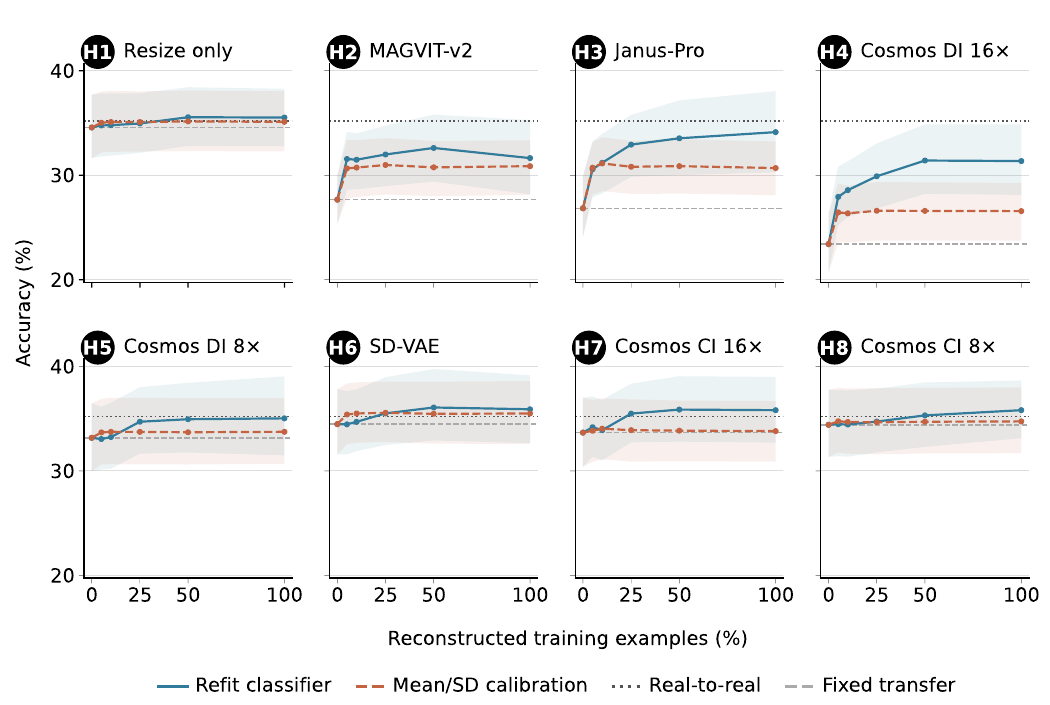}
\caption{\mbox{H1--H8} Touch and Go / DINOv2. Every reconstruction condition, with paired pointwise 95\% cluster-bootstrap intervals. The horizontal references are real-to-real accuracy (dotted) and fixed-probe transfer (grey dashed).}
\label{fig:all-tag-dino}
\includegraphics[width=0.83\textwidth]{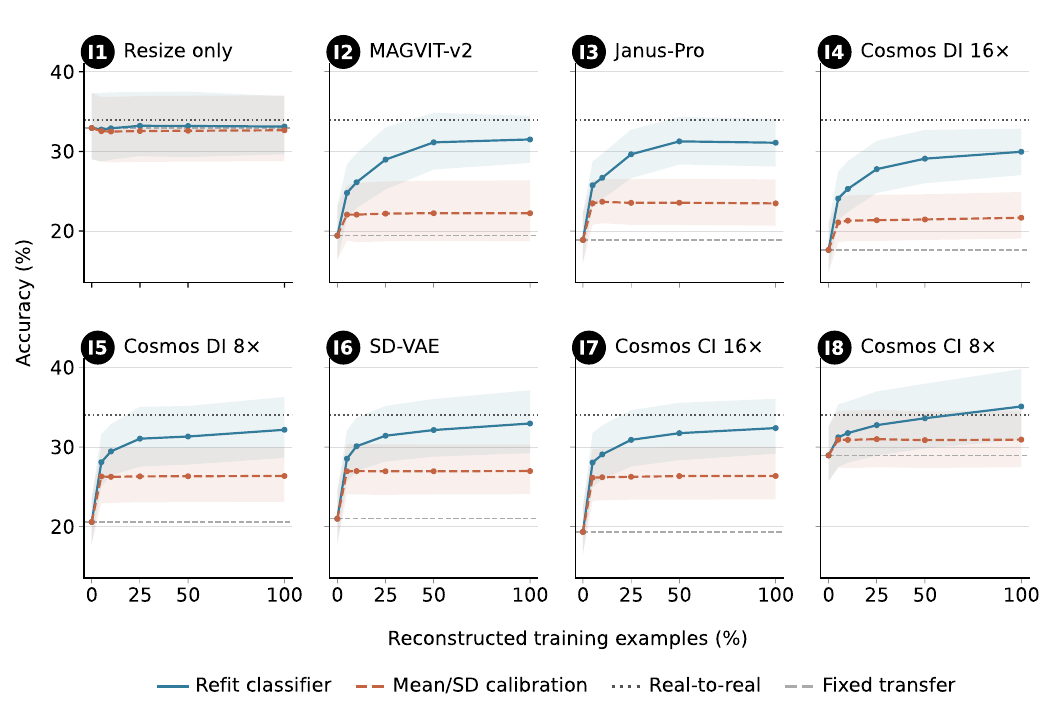}
\caption{\mbox{I1--I8} Touch and Go / CLIP. Every reconstruction condition, with paired pointwise 95\% cluster-bootstrap intervals. The horizontal references are real-to-real accuracy (dotted) and fixed-probe transfer (grey dashed).}
\label{fig:all-tag-clip}
\end{figure*}

Figures~\ref{fig:all-of-real-dino}\,\mbox{F1--F8}, \ref{fig:all-of-real-clip}\,\mbox{G1--G8}, \ref{fig:all-tag-dino}\,\mbox{H1--H8} and \ref{fig:all-tag-clip}\,\mbox{I1--I8} show every released reconstruction condition, including resize-only and the continuous autoencoders. Curves average correctness across three fixed nested-sampling seeds. The fixed-probe and real-to-real references use the same outer folds. Statistical calibration retains the original classifier. Table~\ref{tab:followup-primary} gives simultaneous intervals for the prespecified contrasts.

\end{document}